\documentclass{article}

\usepackage{arxiv}
\RequirePackage{epsfig}
\RequirePackage{amssymb}
\RequirePackage{natbib}
\RequirePackage{graphicx}
\RequirePackage{amsthm}

\usepackage[utf8]{inputenc} 
\usepackage[T1]{fontenc}    
\usepackage{hyperref}       
\usepackage{url}            
\usepackage{booktabs}       
\usepackage{amsfonts}       
\usepackage{nicefrac}       
\usepackage{microtype}      
\usepackage{lipsum}
\usepackage{graphicx}

\usepackage{mathtools}

\usepackage[table,xcdraw]{xcolor}
\usepackage[caption=false,subrefformat=parens,labelformat=parens]{subfig}
\usepackage{verbatim} 
\usepackage{chemformula}
\usepackage{natbib} 
\usepackage{enumitem}
\usepackage{booktabs}
\usepackage{latexsym}
\usepackage{amsmath}
\usepackage{amsfonts}
\usepackage{amssymb}
\usepackage{amsbsy}
\usepackage{amsthm}
\usepackage{bbm}
\usepackage{bm}
\usepackage{caption}
\usepackage{subcaption}
\usepackage{multirow}
\usepackage{siunitx}
\usepackage[ruled]{algorithm2e}
\usepackage{wrapfig}
\usepackage[title,page]{appendix}

\DeclareMathOperator*{\argmin}{argmin}

\usepackage{soul}

\newtheoremstyle{boldremark}
    {\dimexpr\topsep/2\relax} 
    {\dimexpr\topsep/2\relax} 
    {}          
    {}          
    {\bfseries} 
    {.}         
    {.5em}      
    {}          

\theoremstyle{boldremark}

\definecolor{KWgreen}{RGB}{112,173,71} 
\definecolor{KWblue}{RGB}{0,112,192} 
\definecolor{KWred}{RGB}{192,0,0} 
\definecolor{KWpurple}{RGB}{112,48,160} 
\usepackage{colortbl}

\title{A Symbolic and Statistical Learning Framework to Discover Bioprocessing Regulatory Mechanism: \\cell culture example}

\author{Keilung Choy  \\
 Northeastern University
 \And
  Wei Xie \thanks{Corresponding author. Email: w.xie@northeastern.edu} \\
 Northeastern University
\And
  Keqi Wang\\
Northeastern University
 }
\date{}
\begin{document}
\maketitle

\section*{ABSTRACT}



Bioprocess mechanistic modeling is essential for advancing intelligent digital twin representation of biomanufacturing, yet challenges persist due to complex intracellular regulation, stochastic system behavior, and limited experimental data. This paper introduces a symbolic and statistical learning framework to identify key regulatory mechanisms and quantify model uncertainty. Bioprocess dynamics is formulated with stochastic differential equations characterizing intrinsic process variability, with a predefined set of candidate regulatory mechanisms constructed from biological knowledge. A Bayesian learning approach is developed, which is based on a joint learning of kinetic parameters and regulatory structure through a formulation of the mixture model. To enhance computational efficiency, a Metropolis-adjusted Langevin algorithm with adjoint sensitivity analysis is developed for posterior exploration. Compared to state-of-the-art Bayesian inference approaches, the proposed framework achieves improved sample efficiency and robust model selection. An empirical study demonstrates its ability to recover missing regulatory mechanisms and improve model fidelity under data-limited conditions.

\section{INTRODUCTION}
\label{sec:intro}

Over the past several decades, biopharmaceuticals have risen to prominence due to their rapid development and substantial contributions to public health, particularly through the production of vaccines and therapeutics. By 2021, the global market value of biopharmaceuticals reached \$343 billion, with 67\% (107 out of 159) of approved recombinant products manufactured using mammalian cell systems \citep{walsh2022biopharmaceutical}. 
To facilitate intelligent digital twin development for biomanufacturing processes, it is critical to learn the regulatory mechanisms on reaction network dynamics; that means the mechanisms explaining how reaction rates depend on process states such as molecular concentrations, pH level, and temperature. For example, in enzymatic reaction networks, the enzymes could have different structure-function, depending on the environmental conditions, that influences molecule-to-molecule interactions and reaction rates. 
\textit{The proposed symbolic and statistical learning framework for biological system or bioprocess regulatory mechanism learning is general and it can facilitate interpretable and sample efficient learning.} 


In this paper, cell culture will be used for illustration even though the proposed framework is general.
In mammalian cell culture systems, cellular metabolism is governed by a complex network of mechanisms, including feedback inhibition, feedforward activation, and nutrient-sensing pathways \citep{young2013metabolic,yuan2013nutrient}. These {regulatory interactions} are critical in shaping key process outcomes, such as cell growth, productivity, and product quality. Moreover, mammalian cell cultures are inherently sensitive to variations in culture conditions, which can significantly affect yield and critical quality attributes (CQAs) of the final product \citep{dressel2011effects}.
To enhance scientific understanding and improve predictive capabilities, mechanistic dynamic models of mammalian cell culture systems are developed to quantitatively describe cellular behavior and assess product quality attributes. These models integrate established biological mechanisms to provide a system-level representation of cellular regulation, facilitating the analysis of causal interdependencies between process inputs (e.g., nutrient concentrations, dissolved oxygen levels, feeding strategies) and critical outputs (e.g., cell density, product titer, and CQAs such as glycosylation profiles and product integrity). By capturing the dynamic interactions between cellular metabolism and environmental conditions, these models support rational process design, optimization, and control in biopharmaceutical manufacturing.

However, molecular and metabolite heterogeneity 
introduces inherent stochasticity, 
contributing to batch-to-batch variability frequently observed in biomanufacturing \citep{tonn2019stochastic}. This phenomenon, often referred to as double stochasticity, implies that molecular reaction rates are influenced by random fluctuations in state such as species concentrations and environmental conditions. Failure to account for this heterogeneity can result in biased model predictions, underestimation of process variability, and ultimately ineffective and suboptimal control strategies, as the model may overlook critical subpopulation behaviors that impact process robustness and product quality consistency.
To address this issue, stochastic differential equations (SDEs) provide an appropriate framework for modeling cell culture processes, capturing inherent probabilistic nature of these systems. 
The drift and diffusion terms of SDEs can be formulated based on widely applied foundation models 
for enzymatic reactions 
\citep{Sarantos2018}, such as Michaelis–Menten kinetics, representing molecular interactions and explaining bioprocessing regulatory mechanisms. 

Based on the SDE framework, several challenges remain in the construction of {bioprocess regulatory mechanistic models}.
First, the cellular response to environmental perturbations is inherently complex, particularly when accounting for the intricacies of intracellular metabolic networks and their associated regulatory mechanisms. While biological knowledge in this area is relatively well-established, and extensive information on enzymatic reactions is available through literature and public databases such as the BRENDA Enzyme Database \citep{chang2021brenda}, the activation of individual regulatory mechanisms can vary significantly depending on the cell type 
as well as on specific gene expressions, metabolic characteristics, and bioprocess configurations (e.g., batch, fed-batch, and perfusion). Incorporating all potential regulatory mechanisms without discrimination introduces unnecessary complexity. 
Therefore, it is essential to identify and select a parsimonious subset of predefined regulatory mechanisms that are most relevant for a given cell type and culture condition, thereby ensuring biological fidelity while maintaining model simplicity and interpretability.
Second, limited data availability poses a significant challenge for mechanistic model construction 
leading to substantial uncertainty in model estimation on complex regulatory mechanisms. 

To address these challenges, 
a predefined set of candidate regulatory mechanisms is constructed based on literature and public databases. Each candidate model represents a specific combination of active regulatory mechanisms. 
This \textit{ensemble} forms the foundation of a mixture model formulation, where the overall reaction dynamics are expressed as a weighted combination of candidate models, with the weights reflecting the likelihood of each mechanism being active. In addition, this paper introduces a new Bayesian learning approach for model selection and mechanistic parameter inference that explicitly quantifies model estimation uncertainty. Both weights and kinetic parameters are jointly statistically learned to systematically identify the most relevant regulatory mechanisms while accounting for model uncertainty. 
Unlike many existing methods
—such as approximate Bayesian computation (ABC) \citep{sunnaaker2013approximate} and its variants \citep{xie2022sequential}—this framework employs likelihood-based inference. In likelihood-free approaches like ABC, the complexity and stochasticity of cell culture models, as well as very limited data, make generating sufficient sample paths computationally demanding, with low acceptance rates limiting efficiency.

The proposed framework leverages the benefits from \textit{symbolic and statistical learning} to discover missing regulatory mechanisms.
From a symbolic learning perspective, 
the model candidates are constructed based on established scientific understanding of bioprocessing mechanisms, ensuring that the search space is biologically plausible and interpretable. 
From a statistical learning perspective, the proposed Metropolis adjusted Langevin algorithm (MALA) approach adds a drift in Markov chain Monte Carlo (MCMC) posterior sampling search based on the gradient of likelihood that can efficiently allocate more sampling efforts on the most promising regulatory mechanisms, explaining bioprocess dynamics in the observations. Furthermore, adjoint sensitivity analysis, accounting for complex spatial-temporal dependence of candidate models and mechanistic parameters during posterior search, can improve Bayesian learning efficiency and estimation robustness in model selection and parameter inference. 


The structure of this paper is organized as follows. Section~\ref{sec:problemDescription} provides the problem description for bioprocess mechanistic modeling, including a brief introduction to the regulatory mechanisms considered. Leveraging on the information from closed-form posterior distribution, Section~\ref{sec:Meta} describes the MALA posterior inference and asymptotic consistency. Section~\ref{sec:LS-multiScaleModel} presents a new posterior sampling algorithm that utilizes adjoint sensitivity analysis, accounting for interdependencies of model parameters, to accelerate the posterior sampling convergence.
An empirical study is conducted in Section~\ref{sec: empirical study}, demonstrating the framework’s promising performance in terms of 
sample and computational efficiency to discover missing regulatory mechanisms, facilitating the construction of intelligent digital model representations. 
Finally, Section~\ref{sec: conclusion} synthesizes the key findings and insights gathered throughout this study, concluding the paper.

{
}

\section{Problem Description}
\label{sec:problemDescription}

We will provide a rigorous problem description in Section~\ref{subsec:BioprocessRegulatoryModeling} and use a simple representative metabolic reaction network example in Section~\ref{sec:metabolicExample} to explain how the proposed approach builds on pre-defined rules on bioprocessing mechanisms and enables us to leverage the benefits from {symbolic and statistical learning} to discover missing regulatory mechanisms. This can facilitate interpretable and sample-efficient learning.

\subsection{Bioprocess Regulatory Mechanism Modeling}
\label{subsec:BioprocessRegulatoryModeling}

In this paper, bioprocess mechanistic model is represented by stochastic differential equations (SDEs), i.e.,
\begin{equation}\label{evol_cell_conc}
    d\pmb{s}_t = \pmb{\mu}(\pmb{s}_t;\bm{\theta}^c)
    dt+\pmb{\sigma}(\pmb{s}_t;\bm{\theta}^c)dW_t, 
\end{equation}
where $d {W}_t$ is the increment of a standard Brownian motion and $\pmb{s}_t= (s_t^1, s_t^2,\ldots,s_t^p)^\top$ represents a $p$-dimensional state at any time $t$. 
Both mean $\pmb{\mu}(\pmb{s}_t; \pmb{\theta}^c)$ and standard deviation $\pmb{\sigma}(\pmb{s}_t;\bm{\theta}^c)$ are functions of the system state $\pmb{s}_t$ and they are determined by 
unknown regulatory mechanisms characterized by $\mathbf{v}(\pmb{s}_t; \pmb{\theta}^c)$ 
with \( \bm{\theta}^c \) representing the true set of model parameters characterizing regulatory mechanisms.
This SDE-based mechanistic model represents the dynamics and inherent stochasticity of bioprocess, which is driven by fluctuations in enzyme activities, gene expression levels, and environmental conditions.

Suppose the bioprocessing dynamics is induced by a reaction network, composed of $p$ molecular species and $L$ reactions, with structure specified by a known $p \times L$ stoichiometry matrix denoted by $\pmb{N}$.
Let $\bm{R}_t$ be a vector representing the number of occurrences of each molecular reaction within a short time interval $(t, t + \Delta t]$, during which the system state evolves from $\pmb{s}_t$ to $\pmb{s}_{t+1}$. Since a molecular reaction will occur when one molecule {collides, binds, and reacts} with another one while molecules move around randomly, driven by stochastic thermodynamics of Brownian motion \citep{golightly2005bayesian},
the occurrences of molecular reactions 
are modeled by non-homogeneous Poisson process.
Thus, the state transition model 
becomes, 
\begin{equation*}
\pmb{s}_{t+1}=\pmb{s}_{t}+\bm{N}\cdot\bm{R_{t}}
\quad \mbox{with} \quad \bm{R}_{t}\sim \mbox{Poisson}(\bm{v}(\pmb{s}_{t};\bm{\theta}^c)), \label{Poisson}
\end{equation*}
where $\bm{N}\cdot\bm{R_{t}}$ represents the net amount of reaction outputs during time interval $(t,t+\Delta t]$. Then, the bioprocess mechanistic model in Equation~(\ref{evol_cell_conc})
can be 
written in the updated SDE form, i.e.,
\begin{equation}
d\pmb{s}_t =\bm{N}\mathbf{v}(\pmb{s}_t;\bm{\theta}^c)dt+(\bm{N}\mbox{diag}(\mathbf{v}(\pmb{s}_{t};\bm{\theta}^c))\bm{N}^{\top})^{\frac{1}{2}}dW_t.
\label{eq.SDE-model}
\end{equation}

The bioprocess dynamics is specified by a regulatory mechanistic model of reaction network flux rates \( \mathbf{v}(\pmb{s}_t; \pmb{\theta}^c)= \left(v^1(\pmb{s}_t;\bm{\theta}^c), v^2(\pmb{s}_t;\bm{\theta}^c), \dots, v^L(\pmb{s}_t;\bm{\theta}^c)\right)^\top \) for $L$ reactions;   
that depends on state variables $\pmb{s}_t$ such as molecular concentrations, temperature, and pH level. 
However, in real-world applications, the understanding of biological system 
or bioprocessing regulatory mechanism 
is often not fully known since the structure-function 
of biomolecules (such as DNAs, RNAs, and proteins) is very complex and highly depends on various factors, including environmental conditions, the type of substrates, and ion concentrations. Therefore, to construct intelligent digital twins for biomanufacturing systems, this motivates the need to learn the underlying regulatory mechanism, including both structure and parameters of \( \mathbf{v}(\pmb{s}_t; \pmb{\theta}^c) \).

In specific, built on current scientific knowledge, we propose a total of \( K \) candidate models that can come from literature and experimental studies, and each candidate regulation model is denoted as \( \widetilde{\mathbf{v}}_k(\pmb{s}_t; \pmb{\vartheta}_k) \) for \( k = 1,2, \ldots, K \). We assign a weight $w_k$ to each $k$-th candidate model and construct a mixture representation to characterize bioprocess dynamics induced by various potential regulatory mechanisms, 
\begin{equation}
    \widetilde{\mathbf{v}}
    (\pmb{s}_t;\bm{\theta},\bm{w})=\sum_{k=1}^{K} w_k\widetilde{\mathbf{v}}_k(\pmb{s}_t;\pmb{\vartheta}_k) ~\mbox{with}~ 
    ~ w_k\in [0,1] ~\mbox{for}~k=1,2,\ldots,K; ~\mbox{subjected to} ~\sum_{k=1}^{{K}} w_k=1.
    \label{eq.mixV}
\end{equation}
Let $\pmb{\theta} = (\pmb{\vartheta}_1,\pmb{\vartheta}_2,\ldots,\pmb{\vartheta}_K)$ and 
$\pmb{w}=(w_1,w_2,\ldots,w_K)$ with $w_k$ representing the probability that the $k$-th candidate model, denoted by $M_k$, is selected.
Therefore, the candidate model is specified by 
the probabilistic weights and regulatory mechanistic parameters, i.e., $(w_k,\pmb{\vartheta}_k)_{k=1}^K$. 
Denote the set of regulatory mechanisms as $\mathcal{R} = \{R_1, R_2, \ldots, R_C\}$, where $C$ is the total number of regulatory mechanisms of interest; for example in a representative cell culture metabolic reaction network example as shown in Figure~\ref{fig:MetaNetwork} with $C=4$ candidate regulatory mechanisms $R_1,R_2,R_3,R_4$.
The total number of model candidates, corresponding to all possible combinations of activation statuses (i.e., active or inactive) of each regulatory mechanism, 
could become $K = 2^C$.



In this paper, we propose an efficient and interpretable Bayesian learning approach that can quickly identify missing regulatory mechanisms and advance scientific understanding. Let $\{\widetilde{\pmb{s}}_t\}$ denote the prediction on the trajectory $\{\pmb{s}_t\}$ generated by underlying regulatory mechanism \( \mathbf{v}(\pmb{s}_t; \pmb{\theta}^c) \) by using a candidate model of $\widetilde{\mathbf{v}}(\pmb{s}_t;\bm{\theta},\bm{w})$ with a mixture form as shown in Equation~(\ref{eq.mixV}). Given any posterior sample of the regulatory model specified by $(\pmb{\theta},\pmb{w})$, 
the gradient of the log-likelihood or posterior with respect to both $\bm{\theta}$ and $\bm{w}$ is derived, enabling the use of Metropolis-adjusted Langevin Algorithm (MALA) to drift posterior sampling to the area of $(\pmb{\theta},\pmb{w})$ with high likelihood. 
To further accelerate the convergence of MALA, adjoint sensitivity analysis (SA) is applied to quantify how the posterior state evolves with respect to the initial sample and then a metamodel is constructed to estimate the initial bias, reduce the warmup, and
guide the posterior sampling toward high-probability regions of the candidate model space. 

Therefore, based on candidate models from existing scientific knowledge characterizing molecule-to-molecule interactions and the potential logic of reaction network dynamics,
the proposed Bayesian learning approach combines likelihood-gradient-driven posterior sampling and adjoint SA-based metamodel correction, ensuring interpretable, efficient, and robust learning of the underlying regulatory mechanisms.

\subsection{A Representative Metabolic Reaction Network Illustrative Example
}
\label{sec:metabolicExample}

\begin{figure}[htb!]
    \centering\includegraphics[width=1\textwidth]{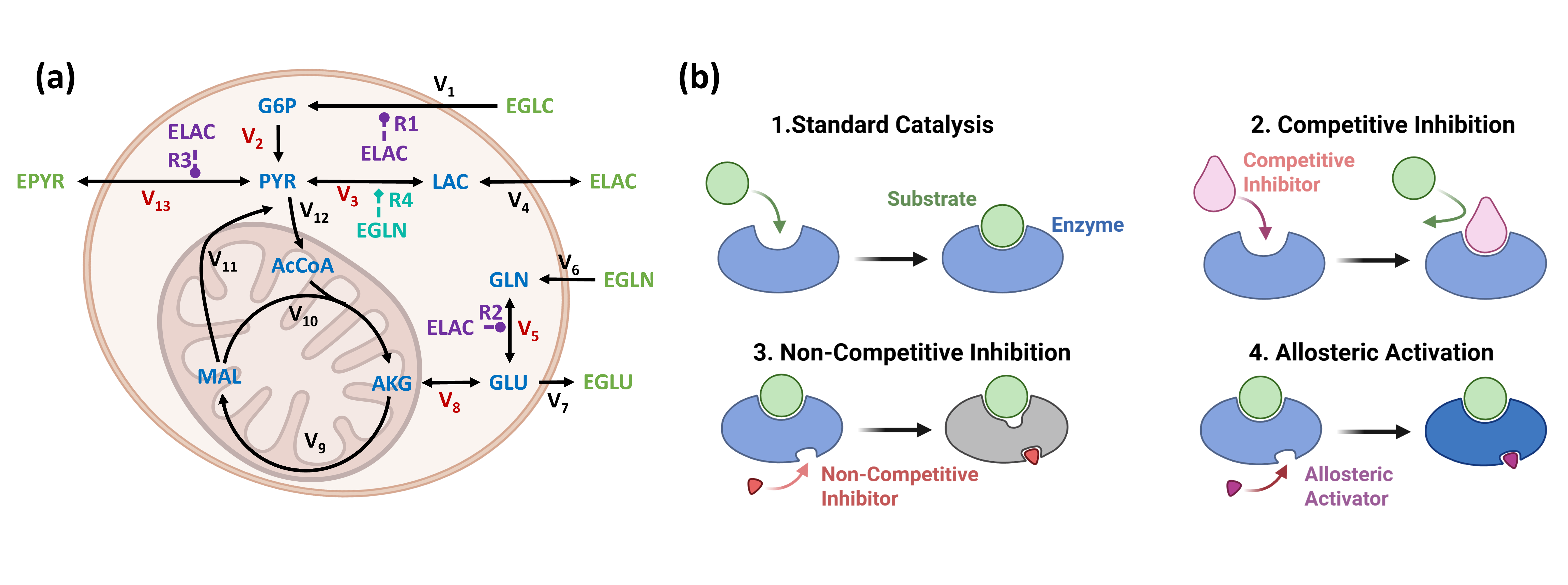}
    \vspace{-0.35 in}
    \caption{(a) Schematic of a simple metabolic reaction network with \textcolor{KWgreen}{green} and \textcolor{KWblue}{blue} represent extracellular and intracellular metabolites.
    Reactions with flux rate modeling using M-M kinetics are shown in  \textcolor{KWred}{Red}. 
    (b) \textit{Illustration of enzyme regulatory mechanisms} (Created with BioRender.com).
    (1) Standard Catalysis: Under baseline conditions, the substrate binds to the enzyme's active site, leading to catalysis without regulatory interference.
    (2) Competitive Inhibition: A competitive inhibitor binds to the enzyme's active site, preventing substrate binding and thus inhibiting catalysis.
    (3) Non-Competitive Inhibition: A non-competitive inhibitor binds to a distinct allosteric site, allowing substrate binding but impairing catalytic activity, resulting in reduced reaction rates without affecting substrate affinity.
    (4) Allosteric Activation: An allosteric activator binds to an allosteric site, inducing a conformational change that enhances enzyme activity by either improving substrate binding affinity, increasing catalytic turnover, or both.
    }
    \label{fig:MetaNetwork}
\end{figure}

In this section, we describe the kinetic modeling of regulatory metabolic networks for a simple representative cell culture example as shown in Figure~\ref{fig:MetaNetwork}. The kinetic model in Equation~(\ref{eq.mixV}) has the selected components providing the \textit{logic representation} of the regulatory mechanisms of the cellular reaction network system. This enables us to learn the underlying mechanisms explaining the dynamics of extracellular and intracellular metabolite concentrations. The reaction rate and the change of metabolites depends on substrate availability and regulatory interactions of molecules. Therefore, the kinetic model of the metabolic reaction network characterizes the underlying mechanisms governing cellular metabolic processes and their dynamic responses to environmental changes.

For each $\ell$-th enzymatic regulation in the metabolic reaction network with structure specified by a stoichiometry matrix $\pmb{N}$, its reaction or flux rate at time $t$ is modeled as below, following the Michaelis–Menten (MM) kinetics \citep{MM2007}:
\begin{equation}
    v^\ell(\pmb{s}_t;\bm{\theta}) = V_{\max,\ell} \prod_{y \in \Omega_Y^\ell} \frac{s_t^y}{s_t^y + K_{m,y}},\label{metabolic_flux}
\end{equation}
where the set $\Omega_Y^\ell$ represents the collection of substrates influencing the flux rates. The parameters $K_{m,y}$ and $V_{\max,\ell}$ represent the affinity constant and the maximum specific flux rate, respectively.
In specific, for each enzymatic reaction, 
$E+S\underset{k_R}{\overset{k_F}{\rightleftarrows}} ES \overset{k_{cat}}{\rightarrow} E+P~\text{(product)}$, the substrate ($S$) needs to interact and form a reversible complex ($ES$) with the enzyme ($E$)
for the enzyme to be able to perform its catalytic function to produce the product ($P$). 
\textit{Kinetic rates} include: (1) $k_F$ and $k_R$ associated with the binding and unbinding rates of molecules $E$ and $S$; and (2) $k_{cat}$ reflects the enzyme's efficiency in facilitating molecular reactions by reducing the required energy barrier. 
For example, the MM kinetics has the key building blocks with the form of $
\frac{V_{max}{s}^y_t}{{s}^y_t+K_{m,y}}$ as shown in Equation~(\ref{metabolic_flux}),
including the mechanistic parameters characterizing the regulatory mechanisms of the enzymatic reaction: 
(1) $V_{max}=k_{cat}[E]_{total}$ is the maximum possible velocity of the molecular reaction that can occur when all the enzyme molecules are bound with certain specific substrate; 
and (2) $K_{m}=\frac{k_R+k_{cat}}{k_F}=\frac{[E][S]}{[ES]}$ is a dissociation constant measuring the affinity of $E$ for $S$
and a low value of $K_{m}$ is often interpreted as a high affinity of the enzyme for the substrate.


Various regulatory mechanisms—including allosteric regulation, competitive inhibition, and non-competitive inhibition—are incorporated into the proposed metabolic flux kinetic model (Equation~\ref{metabolic_flux}). A brief introduction to these common mechanisms is provided in the caption of Figure~\ref{fig:MetaNetwork}, along with an illustrative diagram.
Considering multiple regulatory mechanisms, the flux rate model for each $\ell$-th reaction is updated as follows:
\begin{align*}
    v^\ell(\pmb{s}_t;\bm{\theta}) = v_{\max,\ell} \prod_{y \in \Omega_Y^\ell} \frac{s_t^y}{s_t^y + K^{'}_{m,y}}\prod_{z \in \Omega_Z^\ell} \frac{K_{i,z}}{s_t^z + K_{i,z}}
    ~~~\mbox{with} ~~~
    K^{'}_{m,y}=K_{m,y}\Big(1 + \sum_{z' \in \Omega_{Z'}^\ell}\frac{s_t^{z'}}{K_{i,z'}} + \sum_{x \in \Omega_{X}^\ell}\frac{K_{a,x}}{s_t^{x}}\Big),\nonumber
\end{align*}
where $\Omega_Y^\ell$ denotes the set of substrates, $\Omega_Z^\ell$ the set of non-competitive inhibitors, $\Omega_{Z'}^\ell$ the set of competitive inhibitors, and $\Omega_X^\ell$ the set of allosteric activators.
$K_{i}$ and $K_{a}$ are the inhibition and activation constants.

A representative metabolic reaction network for mammalian cells, adapted from the studies, including \cite{hassell1991growth,mulukutla2012metabolic,ghorbaniaghdam2014analyzing,ghorbaniaghdam2014silico,wang2024metabolic,wang2024multi}, is illustrated in Figure~\ref{fig:MetaNetwork}. The stoichiometry of all relevant reactions in CHO cell cultures is detailed in Table~\ref{tab:reaction}. Reactions highlighted in red in Figure~\ref{fig:MetaNetwork} indicate those for which a regulation model has been constructed, incorporating the key regulatory mechanisms: (1) non-competitive inhibition, where a metabolite indirectly reduces enzyme activity by binding to a regulatory site (e.g., \textbf{R1} and \textbf{R2}); (2) competitive inhibition, where metabolites compete with the substrate for the active site (e.g., \textbf{R3}); and (3) allosteric activation, where a metabolite enhances enzyme activity by binding to an allosteric site (e.g., \textbf{R4}). For the remaining reactions, shown in black, the pseudo-steady-state assumption is applied. The comprehensive reaction rate model is provided in Table~\ref{tab:kinetic}. This example will be used in the empirical study to demonstrate the performance of the proposed symbolic and Bayesian learning approach.

\section{Bayesian Network Inference and Regulatory Mechanism Discovery}
\label{sec:Meta}

We first derive the posterior inference for the mechanistic model Bayesian learning in Section~\ref{subsec:BayesianInference} and then conduct asymptotic study in Section~\ref{subsec:asymptoticStudy} to show that it can recover missing regulatory mechanisms.

\subsection{Bayesian Inference to Improve Model Prediction}
\label{subsec:BayesianInference}

Our purpose is to efficiently learn the missing regulatory mechanism. 
Since the posterior inference of the mechanistic model (\ref{eq.mixV}) is typically complex and intractable, 
we employ the MALA 
\citep{roberts1996exponential} that utilizes the likelihood gradient to guide the search for the true mechanism $\mathbf{v}(\pmb{s}_t; \pmb{\theta}^c)$
through drifting the MCMC sampling of $(\bm{\theta}, \bm{w})$ 
toward regions with high likelihood; this can improve computational efficiency and convergence speed compared to traditional MCMC with random-walk proposals. In specific, 
the Langevin diffusion process of MALA is represented below,
\begin{equation}
\label{cont_MALA}
\begin{aligned}
    d\bm{\theta}_{\tau} = \frac{1}{2} \nabla_{\bm{\theta}} \log P(\bm{\theta}_{\tau}, \bm{w}_{\tau} \mid \mathcal{D}_m) \, d\tau + dW_{\tau} ~\mbox{and}~
    d\bm{w}_{\tau} = \frac{1}{2} \nabla_{\bm{w}} \log P(\bm{\theta}_{\tau}, \bm{w}_{\tau}  \mid \mathcal{D}_m) \, d\tau + dW_{\tau}.
\end{aligned}
\end{equation}
As sampling proceeds with $\tau \to \infty$, the distribution of $\pmb{\theta}_{\tau}$ converges to the desired posterior distribution. Then, through discretization with step size  $\epsilon$, we have the MALA sampling update at each $\tau$-th iteration,
\begin{equation}
\label{disc_MALA}
\left\{
\begin{aligned}
    \bm{\theta}_{\tau+1} &= \bm{\theta}_{\tau} + \frac{\epsilon^2}{2} \nabla_{\bm{\theta}} \log P( \bm{\theta}_{\tau}, \bm{w}_{\tau} \mid \mathcal{D}_m) + \epsilon \cdot \mathbf{z}_{\theta}, \quad \mathbf{z}_{\theta} \sim \mathcal{N}(0, I), \\
    \bm{w}_{\tau+1} &= \bm{w}_{\tau} + \frac{\epsilon^2}{2} \nabla_{\bm{w}} \log P(\bm{\theta}_{\tau}, \bm{w}_{\tau} \mid \mathcal{D}_m) + \epsilon \cdot \mathbf{z}_{w}, \quad \mathbf{z}_{w} \sim \mathcal{N}(0, I).
\end{aligned}
\right.
\end{equation}

We first calculate the likelihood $P(\mathcal{D}_m \mid \bm{\theta}, \bm{w})$. 
Given the candidate model $\widetilde{\mathbf{v}}(\pmb{s}_t; \bm{\theta}, \bm{w})$ in Equation~(\ref{eq.mixV}), we can generate bioprocess state trajectories, denoted by $\{ \widetilde{\pmb{s}}_t \}$ using the SDE, i.e.,
\begin{align}
    d\widetilde{\pmb{s}}_{t} = \bm{N} \widetilde{\mathbf{v}}(\widetilde{\pmb{s}}_{t};\bm{\theta},\bm{w})dt+\left[\bm{N}\mbox{diag}(\widetilde{\mathbf{v}}(\widetilde{\pmb{s}}_{t};\bm{\theta},\bm{w}))\bm{N}^{\top}\right]^{\frac{1}{2}}d\widetilde{W}_t,\label{simulator}
\end{align}
where $d\widetilde{W}_t$ is the increment of a standard Brownian motion.
Let $\mathcal{D}_m=\{ \boldsymbol{\zeta}^i \}_{i=1}^{m}$ represent $m$ independent trajectory observations obtained from the real system. Each trajectory consists of $H$ state-transition observations, i.e.,
$
\boldsymbol{\zeta}^i = \{ \pmb{s}^i_{t_1}, \pmb{s}^i_{t_2}, \ldots, \pmb{s}^i_{t_H}, \pmb{s}^i_{t_{H+1}} \}$. To facilitate learning of the missing mechanisms from data $\mathcal{D}_m$, 
we first derive the closed-form likelihood, i.e.,  
$
    P(\mathcal{D}_m|\bm{\theta},\bm{w})=\prod_{i=1}^{m}  P\left( \pmb{s}_{t_1}^i\right) \prod_{h=1}^{H} P(\left. \pmb{s}_{t_{h+1}}^i\right| \pmb{s}^i_{t_h} ; \bm{\theta},\bm{w} ) 
$.

For notation simplification, suppose that data collection frequent is fixed with $\Delta t=t_{h+1}-t_h$ for $h=1,2,\ldots,H$. 
Based on (\ref{simulator}), we can approximate the conditional distribution  
$P( \pmb{s}_{t_{h+1}}^i\mid \pmb{s}^i_{t_h} ; \bm{\theta},\bm{w})$
as $ \mathcal{N} ( \bm{\mu}_{t_h}^i, \Sigma_{t_h}^i )$ with
$\bm{\mu}_{t_h}^i :=\pmb{s}_{t_h}^i + \bm{N} \widetilde{\mathbf{v}}(\pmb{s}_{t_h}^i; \bm{\theta},\bm{w}) \Delta t$ and $\Sigma_{t_h}^i :=\bm{N}\mbox{diag}(\widetilde{\mathbf{v}}(\pmb{s}_{t_h}^i;\bm{\theta},\bm{w}))\bm{N}^{\top} \Delta t.$
With this approximation, we further derive the posterior, 
$
P(\bm{\theta},\bm{w} \mid \mathcal{D}_m) \propto P(\bm{\theta}) P(\bm{w}) P(\mathcal{D}_m|\bm{\theta},\bm{w})= P(\bm{\theta})P(\bm{w}) \prod_{i=1}^{m} \left[ P\left(\pmb{s}_{t_1}^i\right) \prod_{h=1}^{H} P\left( \pmb{s}_{t_{h+1}}^i\mid \pmb{s}^i_{t_h} ; \bm{\theta},\bm{w} \right) \right],
\nonumber
$
and then take log transformation,
\begin{equation}
\log P(\bm{\theta}, \bm{w} \mid \mathcal{D}_m) 
 = \log P(\bm{\theta}) + \log P(\bm{w})-\frac{1}{2}  \sum_{i=1}^m \sum_{h=1}^H\left[
2\pi\log |\Sigma_{t_h}^i |
+ \left( \pmb{s}_{t_{h+1}}^i - \bm{\mu}_{t_h}^i \right)^\top {\Sigma_{t_h}^i}^{-1} \left( \pmb{s}_{t_{h+1}}^i - \bm{\mu}_{t_h}^i \right)
\right].
\label{eq.posterior}
\end{equation}

To further employ MALA in (\ref{disc_MALA}), we calculate the gradients of the log-posterior with respect to $(\pmb{\theta}, \pmb{w})$, 
\begin{eqnarray}
\lefteqn{\nabla_{\bm{\theta}} \log P(\bm{\theta}, \bm{w} \mid \mathcal{D}_m) 
= \nabla_{\bm{\theta}} \log P(\bm{\theta}) 
- 
\frac{1}{2} \sum_{i=1}^m \sum_{h=1}^H 
\left[
\operatorname{Tr} \left( (\Sigma_{t_h}^i)^{-1} \nabla_{\bm{\theta}} \Sigma_{t_h}^i \right) \right.}
\nonumber \\
&& \left. 
- 2 \left( \nabla_{\bm{\theta}} \bm{\mu}_{t_h}^i \right)^\top (\Sigma_{t_h}^i)^{-1} (\pmb{s}_{t_{h+1}}^i - \bm{\mu}_{t_h}^i) + (\pmb{s}_{t_{h+1}}^i - \bm{\mu}_{t_h}^d)^\top (\Sigma_{t_h}^i)^{-1} \left( \nabla_{\bm{\theta}} \Sigma_{t_h}^i \right) (\Sigma_{t_h}^i)^{-1} (\pmb{s}_{t_{h+1}}^i - \bm{\mu}_{t_h}^i)\right],
\nonumber 
\end{eqnarray}
and
\begin{eqnarray}
\lefteqn{\nabla_{\bm{w}} \log P(\bm{\theta}, \bm{w} \mid \mathcal{D}_m) 
= \nabla_{\bm{w}} \log P(\bm{w}) 
- \frac{1}{2} \sum_{i=1}^m \sum_{h=1}^H 
\left[
\operatorname{Tr} \left( (\Sigma_{t_h}^i)^{-1} \nabla_{\bm{w}} \Sigma_{t_h}^i \right) \right.}
\nonumber \\
&& \left.
- 2 \left( \nabla_{\bm{w}} \bm{\mu}_{t_h}^i \right)^\top (\Sigma_{t_h}^i)^{-1} (\pmb{s}_{t_{h+1}}^i - \bm{\mu}_{t_h}^i) 
+ (\pmb{s}_{t_{h+1}}^i - \bm{\mu}_{t_h}^i)^\top (\Sigma_{t_h}^i)^{-1} \left( \nabla_{\bm{w}} \Sigma_{t_h}^i \right) (\Sigma_{t_h}^i)^{-1} (\pmb{s}_{t_{h+1}}^i - \bm{\mu}_{t_h}^i)
\right].
\nonumber
\end{eqnarray}

\subsection{Asymptotic Study} 
\label{subsec:asymptoticStudy}
Suppose the true model $\mathbf{v}(\pmb{s}_t; \bm{\theta}^c)$ is included in the set of candidate models; that means there exists some index $k^*$ such that $\mathbf{v}(\pmb{s}_t; \bm{\theta}^c) = \widetilde{\mathbf{v}}_{k^*}(\pmb{s}_t; \bm{\theta}^c)$. We define the true weight vector $\bm{w}^c$, which consists of zeros except for $w_{k^*} = 1$. Under this setting, we can show that the proposed approach achieves asymptotic convergence, meaning that the posterior probabilities, i.e., $P(\bm{w} = \bm{w}^c\mid \mathcal{D}_m)$ and  $P(\bm{\vartheta}_{k^*} = \bm{\theta}^c \mid \mathcal{D}_m)$, both converge to 1  as $m\rightarrow \infty$.
To establish this consistency result, we first define
$\mathcal{W} := \left\{ \bm{w} \in [0,1]^K : \sum_{k=1}^K w_k = 1 \right\} \subseteq \mathbb{R}^K,$
which represents the space of $\bm{w}$. Let $\Theta_k$ denote the space of $\pmb{\vartheta}_k$ for $k=1,2,\ldots,K$. We define $\Theta$ as the union of all parameter spaces, i.e., $\Theta = \bigcup_{k=1}^K \Theta_k$. 
We define a metric denoted by
$d_{\Theta_k}(\pmb{\vartheta}_{k}, \pmb{\vartheta}_{k}') = \|\pmb{\vartheta}_{k}, \pmb{\vartheta}_{k}'\|,$ and $d_{\mathcal{W}}(\bm{w}, \bm{w}') = \|\bm{w}, \bm{w}'\|,$
where $\|\cdot\|$ denotes the Euclidean norm, and further define an open ball
$
{B}(\pmb{\vartheta}_{k}, \delta) = \left\{ \pmb{\vartheta}_k' \in \Theta_k : d_{\Theta_k}(\pmb{\vartheta}_{k}', \pmb{\vartheta}_{k}) < \delta \right\}$ and $
{B}(\pmb{w}, \delta) = \left\{ \bm{w}' \in \mathcal{W} : d_{\mathcal{W}}(\bm{w}',\bm{w}) < \delta \right\}.$

Given the observations $\{ \boldsymbol{\zeta}^i \}_{i=1}^{m}$, 
we have the asymptotic consistency of Bayesian inference in Theorem 1. 

\textbf{Theorem 1}\citep{Miller2023} Assume $\widetilde{\mathbf{v}}(\pmb{s}_t; \bm{\theta},\bm{w})$ is specified by $(\bm{w},\pmb{\theta})$. 
Suppose $\bm{w}^c \in \mathcal{W}$ and $\bm{\theta}^c \in \Theta_{k*}$. If $\boldsymbol{\zeta}^1, \boldsymbol{\zeta}^2, \ldots,\boldsymbol{\zeta}^m$ are i.i.d. and follow the SDE in (\ref{evol_cell_conc}), then for any $
\delta 
> 0$, we have the asymptotic consistency, 
\begin{equation*}\label{theorem}
\lim_{m \to \infty} P(\bm{w} \in {B}(\bm{w} ^c, \delta)  \mid \boldsymbol{\zeta}^1, \boldsymbol{\zeta}^2, \ldots,\boldsymbol{\zeta}^m) = 1
~\mbox{and}~
\lim_{m \to \infty} P(\bm{\vartheta}_{k^*} \in {B}(\bm{\theta} ^c, \delta)  \mid \boldsymbol{\zeta}^1, \boldsymbol{\zeta}^2, \ldots,\boldsymbol{\zeta}^m) = 1 ~ \text{a.s.} 
\end{equation*}

To complete the consistency result, it is necessary to establish the uniform convergence of the flux rate model. Specifically, since $\widetilde{\mathbf{v}}(\pmb{s}; \bm{\theta}, \bm{w})$ is continuous with respect to $(\bm{\theta}, \bm{w})$, 
by applying the Continuous Mapping Theorem, we have the regulatory model estimate $\widetilde{\mathbf{v}}(\pmb{s}; \bm{\theta}, \bm{w})$ converges to $
\mathbf{v}(\pmb{s}; \bm{\theta}^c)$ in probability. 

\section{MALA Posterior Sampling with Adjoint Sensitivity Analysis
}
\label{sec:LS-multiScaleModel}


When MALA is used to generate Bayesian posterior samples following Equation~(\ref{disc_MALA}), warm-up process is often computationally expensive and time-consuming. 
To accelerate the convergence of posterior sampling process,
we exploit the MALA dynamics
and calculate the sensitivity of the ``steady-state" 
$(\bm{\theta}_{T},\bm{w}_{T})$ for any large $T$, representing the solution of (\ref{cont_MALA}), with respect to the initial sample $( \bm{\theta}_0,\bm{w}_0)$, defined as
$
\bm{J}_{0,T}(\bm{\theta}_0,\bm{w}_0) := \left[ \frac{\partial (\bm{\theta}_{T},\bm{w}_{T})}{\partial (\bm{\theta}_0,\bm{w}_0)}\right]
$
that enables us to quickly estimate the initial bias needed to remove. 

\textit{In specific, the adjoint sensitivity analysis on the SDE (\ref{cont_MALA}), accounting for spatial-temporal interdependence, 
is utilized to efficiently estimate the initial bias through calculating $\mathbb{E}[\bm{J}_{0,T}]$ and local metamodeling.} 
Suppose (1) $\widetilde{\mathbf{v}}(\pmb{s}_t;\bm{\theta},\bm{w})$ is 
infinitely differentiable and bounded related to $(\bm{\theta},\bm{w})$; then $\bm{\mu}_{t_h}^i,\Sigma_{t_h}^i \in C_b^{\infty,1}$; and (2) the first-order derivatives $ \frac{\partial \widetilde{\mathbf{v}}}{\partial \bm{\theta}}$ and $\frac{\partial \widetilde{\mathbf{v}}}{\partial \bm{w}}$ are bounded. Consequently, the posterior in (\ref{eq.posterior}) is also infinite differentiability, i.e., $\nabla_{\bm{\theta}} \log P( \bm{\theta}_{\tau}, \bm{w}_{\tau} \mid \mathcal{D}_m)\in C_b^{\infty,1}$. Thus, given any initial point $(\bm{\theta}_0,\bm{w}_0)$, a unique solution to the SDE (\ref{cont_MALA}) is guaranteed to exist. We use $\Phi_{0,T}(\bm{\theta}_0,\bm{w}_0)$ to represent the solution of the SDEs 
in (\ref{cont_MALA}) at $T$ and it is 
called forward flow satisfying the property:
$
     \Phi_{0,T}(\bm{\theta}_0,\bm{w}_0)=\Phi_{\tau,T}(\Phi_{0,\tau}(\bm{\theta}_0,\bm{w}_0)) ~\mbox{ for } ~0\leq \tau\leq T.
$

We could also generate the inverse flow $\psi_{0,T}:=\Phi_{0,T}^{-1}$ from the SDE system in (\ref{cont_MALA}) \citep{kunita2019stochastic}, 
\begin{equation}
\label{cont_MALA_inv}
\begin{aligned}
    d\bm{\theta}_{\tau} = -\frac{1}{2} \nabla_{\bm{\theta}} \log P(\bm{\theta}_{\tau}, \bm{w}_{\tau} \mid \mathcal{D}_m) \, d\tau - d\widetilde{W}_{\tau} ~~\mbox{and}~~
    d\bm{w}_{\tau} = -\frac{1}{2} \nabla_{\bm{w}} \log P(\bm{\theta}_{\tau}, \bm{w}_{\tau}  \mid \mathcal{D}_m) \, d\tau - d\widetilde{W}_{\tau},
\end{aligned}
\end{equation}
where $\widetilde{W}_{\tau}$ is the backward Wiener process defined as $\widetilde{W}_{\tau}=W_{\tau}-W_T$ for any $ \tau<T$.
With $\Phi_{0,T}$ and $\psi_{0,T}$, we can derive the Jacobian $\bm{J}_{0,T}$ 
\citep{choy2024adjoint}. Define
$
A_{0,T}(\bm{\theta}_T,\bm{w}_T) := \nabla \Phi_{0,T} \big( \psi_{0,T}( \bm{\theta}_T,\bm{w}_T) \big);
$
then,
\begin{equation}\label{sens_inv}
A_{0,T}(\bm{\theta}_T,\bm{w}_T) = \mathbb{I}_d + \frac{1}{2} \int_0^{T}
\begin{bmatrix}
\frac{\partial^2 \log P(\psi_{\tau,T}(\bm{\theta}_T,\bm{w}_T) \mid \mathcal{D}_m)}{\partial \bm{\theta}^2} \\
\frac{\partial^2 \log P(\psi_{\tau,T}(\bm{\theta}_T,\bm{w}_T) \mid \mathcal{D}_m)}{\partial \bm{w}^2}
\end{bmatrix}
A_{\tau,T}(\bm{\theta}_T,\bm{w}_T) \, d\tau.
\end{equation}
Since the diffusion term in (\ref{cont_MALA}) is constant, it vanishes when computing the first-order derivative. By following our previous study \citep{choy2024adjoint}, we have 
$
\bm{J}_{0,T} = A_{0,T} \big( \Phi_{0,T}(\bm{\theta}_0,\bm{w}_0) \big),
$
accounting for the structural information in the SDE of MALA dynamics through 
forward and backward propagation. 
The expected sensitivity
$
\mathbb{E}[\bm{J}_{0,T}]
$
is further used to estimate initial bias and accelerate the convergence of the posterior sampling process.


\label{sec:algorithm}

\begin{algorithm}[H]
\DontPrintSemicolon
\SetAlgoLined
\KwIn{Prior distributions $\pi(\bm{\theta})$ and $\pi(\bm{w})$; step size $\epsilon > 0$; number of samples ${n}$ required for estimation of $\mathbb{E}[\bm{J}_{0,T}]$; number of metamodel training samples $G_{meta}$; total number of required posterior samples $G$; 
warmup length $T$; integer $\Delta$ to reduce sample correlation;
observed data $\mathcal{D}_m = \{ \boldsymbol{\zeta}^i \}_{i=1}^m$; initialize $g'=1$ and $g = 1$.}
\KwOut{Posterior samples $\{(\bm{\theta}^{(g)}_T, \bm{w}^{(g)}_T)\}$ with $g=1,2,\ldots,G$}

\For {$g'=1$ to $G_{meta}$}{
  \textbf{1.} Sample initial parameters: $\bm{\theta}^{(g')}_0 \sim \pi(\bm{\theta})$, $\bm{w}^{(g')}_0 \sim \pi(\bm{w})$.\;

  \For {$j=1$ to $n$}{
    \For{$\tau = 0$ \KwTo $T - 1$}{
      \textbf{2.} Generate $(\bm{\theta}^{(g',j)}_{\tau+1}, \bm{w}^{(g',j)}_{\tau+1})$ using (\ref{disc_MALA}) and then project $\bm{w}^{(g',j)}_{\tau+1}$ to feasible set $\mathcal{W}$.\;

      \textbf{3.} Compute the Metropolis acceptance probability $\alpha_\tau$ and accept/reject the proposal.\;
    }

    \For{$\tau = T-1$ \KwTo $0$}{
      \textbf{4.} Compute the inverse flow $\psi_{\tau,T}(\bm{\theta}^{(g',j)}_T,\bm{w}^{(g',j)}_T)$ using (\ref{cont_MALA_inv}).\;

      \textbf{5.} Compute $\bm{J}_{\tau,T}^{(j)}(\bm{\theta}^{(g')}_0,\bm{w}^{(g')}_0)=A_{\tau,T}(\bm{\theta}^{(g',j)}_T,\bm{w}^{(g',j)}_T)$ via (\ref{sens_inv}).\;
    }
  }

  \textbf{6.} Estimate $\mathbb{E}[\bm{J}_{0,T}(\bm{\theta}^{(g')}_0,\bm{w}^{(g')}_0)]$ with $ \frac{1}{n} \sum_{j=1}^n\bm{J}_{0,T}^{(j)}(\bm{\theta}^{(g')}_0,\bm{w}^{(g')}_0)$. \;
}

\For {$g=1$ to $\left \lceil{G/B}\right \rceil$}{
  \textbf{7.} Sample initial parameters: $\bm{\theta}^{(g)}_0 \sim \pi(\bm{\theta})$, $\bm{w}^{(g)}_0 \sim \pi(\bm{w})$.\;

  \textbf{8.} Predict $(\bm{\theta}^{(g)}_T, \bm{w}^{(g)}_T)$ using the metamodel trained on $\{(\bm{\theta}_0^{(g')}, \bm{w}_0^{(g')}), \bm{J}_{0,T}^{(g')}\}_{g'=1}^{G_{meta}}$.\;
\textbf{9.} Generate $(\bm{\theta}^{(g)}_{T+b\Delta}, \bm{w}^{(g)}_{T+b\Delta})$ using (\ref{disc_MALA}) for $b=1,2,\ldots, B$ and project $\bm{w}^{(g)}_{T+b\Delta}$ to feasible set $\mathcal{W}$. Record posterior samples for every $\Delta$ steps to reduce time serious dependence.\;
}
\caption{Adjoint SA Accelerated MALA for Regulatory Mechanism Learning}
\label{Algr:adjointSA-MALA}  
\end{algorithm}

The adjoint sensitivity analysis assisted MALA presented in Algorithm~\ref{Algr:adjointSA-MALA} proceeds in two stages. In Steps~1-6, we first build a local metamodel to approximate $\mathbb{E}[\bm{J}_{0,T}]$ and estimate the initial bias. 
For each training sample, it begins by drawing  \((\bm{\theta}_0, \bm{w}_0)\) from the priors, simulates forward using a discretized MALA scheme (\ref{disc_MALA}),
and collects trajectories up to \(T\). 
Next, it computes the inverse flow backward from \(\tau=T\) to \(0\) to recover how $(\bm{\theta}_{\tau}, \bm{w}_{\tau})$ changes propagate through the MALA posterior sampling process, derives the pathwise adjoint sensitivity matrix \(\bm{J}_{0,T}(\bm{\theta}_0, \bm{w}_0)\), and estimate $\mathbb{E}[\bm{J}_{0,T}(\bm{\theta}_0, \bm{w}_0)]$. Then, based on Taylor expansion, a local 
metamodel is constructed by identifying the nearest sample (in Euclidean distance) among the metamodel training data and estimate the initial bias for any new initial sample of $(\bm{\theta}_0^{\text{new}},\bm{w}_0^{\text{new}})$, i.e., 
\begin{align*}
    (\bm{\theta}_T^{\text{new}},\bm{w}_T^{\text{new}}) &= (\bm{\theta}_T^{(g^*)},\bm{w}_T^{(g^*)}) + \mathbb{E}[\bm{J}_{0,T}( \bm{\theta}_0^{(g^*)}, \bm{w}_0^{(g^*)})] \cdot \left[ (\bm{\theta}_0^{\text{new}},\bm{w}_0^{\text{new}}) - (\bm{\theta}_0^{(g^*)},\bm{w}_0^{(g^*)}) \right],
\end{align*}
where $g^*=\argmin_{g'\in \{1,\ldots,G_{meta}\}} ||(\bm{\theta}_0^{\text{new}},\bm{w}_0^{\text{new}}),(\bm{\theta}_0^{(g')},\bm{w}_0^{(g')})||$. 
Given multiple space-filling initial points, in Steps~7-9, the metamodel is used to accelerate posterior sampling. For each initial parameters generated from the priors, the metamodel is used to predict the initial bias. These are further refined by continuing the MALA sampling 
with a spacing \(\Delta\) to reduce sample correlation. 
The proposal distribution $q\left(\bm{\theta}^*, \bm{w}^* \mid \bm{\theta}, \bm{w}\right)$ is given 
based on MALA in (\ref{disc_MALA}):
\[
q\left(\bm{\theta}^*, \bm{w}^* \mid \bm{\theta}, \bm{w}\right)
= \mathcal{N}\left(
\bm{\theta}^* \,\middle|\, \bm{\theta} + \frac{\epsilon^2}{2} \nabla_{\bm{\theta}} \log P(\bm{\theta}, \bm{w} \mid \mathcal{D}_m),
\epsilon^2 I
\right)
\times
\mathcal{N}\left(
\bm{w}^* \,\middle|\, \bm{w} + \frac{\epsilon^2}{2} \nabla_{\bm{w}} \log P(\bm{\theta}, \bm{w} \mid \mathcal{D}_m),
\epsilon^2 I
\right).
\]
Metropolis is then used with the acceptance probability,
    $
    \alpha_{\tau} = \min\left\{1, \frac{
    P\left(\left. \bm{\theta}_{\tau+1}, \bm{w}_{\tau+1} \right| \mathcal{D}_m\right) q\left(\left.\bm{\theta}_{\tau}, \bm{w}_{\tau} \right| \bm{\theta}_{\tau+1}, \bm{w}_{\tau+1}\right)
    }{
    P\left(\left. \bm{\theta}_{\tau}, \bm{w}_{\tau} \right| \mathcal{D}_m\right) q\left(\left. \bm{\theta}_{\tau+1}, \bm{w}_{\tau+1}\right|\bm{\theta}_{\tau}, \bm{w}_{\tau}\right)
    }\right\}.
    $

\section{Empirical Study}
\label{sec: empirical study}

To evaluate the proposed framework, we conduct an empirical study using the representative mammalian cell culture system presented in Section~\ref{sec:metabolicExample}. The metabolic reaction network comprises 13 reactions (Table~\ref{tab:reaction}), capturing key pathways within the central carbon network. Among these, five reactions are modeled with explicit regulatory mechanisms (Table~\ref{tab:kinetic}), consistent with the symbolic logic illustrated in Figure~\ref{fig:MetaNetwork}. These include non-competitive inhibition (e.g., ELAC on EGLC uptake in Reaction 2 and on GLN synthetase in Reaction 5), competitive inhibition (e.g., ELAC on EPYR in Reaction 13), and allosteric activation (e.g., EGLN on ELAC production in Reaction 3). These mechanisms correspond to the candidate regulatory modules R1–R4 in Figure~\ref{fig:MetaNetwork}(b), forming the basis for constructing biologically interpretable mixture models.

\begin{table}[htb!]
\centering
\small
\renewcommand{\arraystretch}{1.25}
\caption{The reactions of the cellular metabolic network specify the stoichiometry matrix $\pmb{N}$.}
\label{tab:reaction}
\begin{tabular}{|c|p{6.5cm}|c|p{6.5cm}|}
\hline
\rowcolor[HTML]{C0C0C0} 
\textbf{No.} & \textbf{Reaction} & \textbf{No.} & \textbf{Reaction} \\ \hline
\textbf{1}  & EGLC $\rightarrow$ G6P & \textbf{8}  & GLU $\leftrightarrow$ AKG + NH$_4$ \\ \hline
\textbf{2}  & G6P $\rightarrow$ 2 PYR & \textbf{9}  & AKG $\rightarrow$ MAL + CO$_2$ \\ \hline
\textbf{3}  & PYR $\leftrightarrow$ LAC & \textbf{10} & AcCoA + MAL $\rightarrow$ AKG + CO$_2$ \\ \hline
\textbf{4}  & LAC $\leftrightarrow$ ELAC & \textbf{11} & MAL $\rightarrow$ PYR + CO$_2$ \\ \hline
\textbf{5}  & GLN $\leftrightarrow$ GLU + NH$_4$ & \textbf{12} & PYR $\rightarrow$ AcCoA + CO$_2$ \\ \hline
\textbf{6}  & EGLN $\rightarrow$ GLN & \textbf{13} & EPYR $\rightarrow$ PYR \\ \hline
\textbf{7}  & GLU $\rightarrow$ EGLU & & \\ \hline
\end{tabular}
\end{table}

\begin{table}[hbt!]
\centering
\normalsize
\renewcommand{\arraystretch}{1.8}
\caption{Biokinetic equations for the metabolites fluxes of the model }
\label{tab:kinetic}
\begin{tabular}{|l|p{14cm}|}
\hline
\rowcolor[HTML]{C0C0C0} 
\textbf{No.} & \multicolumn{1}{c|}{\textbf{Pathway with Regulatory Mechanistic Modeling}}                    \\ \hline 
\multicolumn{1}{|l|}{\textbf{2}}   & \large{$v_2 = 
\frac{v_{max, 2} \times EGLC}{K_{m, EGLC}+EGLC} \times \frac{K_{i, ELACtoHK}}{K_{i, ELACtoHK}+ELAC}$}    \\ \hline
\multicolumn{1}{|l|}{\textbf{3}}   & \large{$v_3 = 
\frac{v_{max, 3f}\times EGLC}{K_{m, EGLC} \times (1+\frac{K_{a,EGLN}}{EGLN}) + EGLC} - 
\frac{v_{max, 3r}\times ELAC}{K_{m, ELAC}+ELAC}$}  \\ \hline
\multicolumn{1}{|l|}{\textbf{5}}   & \large{$v_{5} = 
\frac{v_{max, 5f} \times EGLN}{K_{m, EGLN} + EGLN} \times \frac{K_{i, ELACtoGLNS}}{K_{i, ELACtoGLNS}+ELAC} - 
\frac{v_{max, 5r}\times EGLU}{K_{m, EGLU} + EGLU} \times \frac{NH_4}{K_{m, NH_4}+NH_4}$}          \\ \hline
\multicolumn{1}{|l|}{\textbf{8}}   & \large{$v_{8} =  
\frac{v_{max, 8f} \times EGLU}{K_{m, EGLU} + EGLU} - 
\frac{v_{max, 8r} \times NH_4}{K_{m, NH_4}+NH_4}$}                     \\ \hline
\multicolumn{1}{|l|}{\textbf{13}}   & \large{$v_{13} =  
\frac{v_{max, 13} \times EPYR}{K_{m, EPYR} \times (1 + \frac{ELAC}{K_{i,ELACtoPYR}})+ EPYR}$}                  \\ \hline
\end{tabular}
\end{table}

In the simulation setup, all four regulatory mechanisms are assumed to be active in the underlying true model. However, this ground truth is unknown to the learning algorithm during inference; that means the model structures of R1-R4 are known; but we do not know if they are active or not and also the values of their regulatory mechanistic parameters.
To assess the performance of the proposed MALA approach with adjoint sensitivity analysis (SA) and compare it with state-of-the-art Bayesian learning methods for bioprocess mechanistic models, we benchmark against likelihood-free ABC \citep{sisson2018handbook} and standard MALA without adjoint sensitivity \citep{roberts1996exponential}. \textit{The empirical results demonstrate that the proposed Bayesian learning approach performs better in terms of prediction accuracy and recovery of regulatory mechanisms, particularly under data-limited conditions.}

\begin{table}[htbp]
\centering
\caption{The K-S statistics of key states in 72 hours.}
\begin{tabular}{|c|cc|cc|cc|}
\hline
\multirow{2}{*}{State} 
& \multicolumn{2}{c|}{MALA} 
& \multicolumn{2}{c|}{ABC} 
& \multicolumn{2}{c|}{MALA with adjoint sensitivity} \\
\cline{2-7}
& $m=3$ & $m=5$ 
& $m=3$ & $m=5$ 
& $m=3$ & $m=5$ \\
\hline
VCD & $0.42 \pm 0.05$ & $0.40 \pm 0.04$ 
    & $0.44 \pm 0.06$ & $0.41 \pm 0.05$ 
    & $0.35 \pm 0.03$ & $0.33 \pm 0.02$ \\
GLC & $0.37 \pm 0.06$ & $0.35 \pm 0.05$ 
    & $0.39 \pm 0.06$ & $0.36 \pm 0.04$ 
    & $0.31 \pm 0.03$ & $0.29 \pm 0.02$ \\
LAC & $0.48 \pm 0.07$ & $0.46 \pm 0.06$ 
    & $0.50 \pm 0.08$ & $0.47 \pm 0.06$ 
    & $0.39 \pm 0.04$ & $0.37 \pm 0.03$ \\
\hline
\end{tabular}
\label{tab:mala-abc-comparison}
\end{table}

To show the superiority of our proposed approach, we first compare the prediction accuracy of the posterior predictive distribution obtained from MALA with and without adjoint sensitivity analysis, as well as from ABC. Specifically, we generate posterior samples of $(\bm{\theta},\bm{w})$, denoted as $\{(\bm{\theta}^{(g)}, \bm{w}^{(g)})\}_{g=1}^{G}$ and then the sample average approximation (SAA) is used to estimate the posterior predictive distribution 
by evolving the system according to Equation~(\ref{simulator}), i.e.,

\[
P(\pmb{s}_t \mid \pmb{s}_0,\mathcal{D}_m) = \int P( \pmb{s}_t \mid \pmb{s}_0; \bm{\theta}, \bm{w}) \, P(\bm{\theta}, \bm{w} \mid \mathcal{D}_m)  \, d\bm{\theta} d\bm{w}
\approx \frac{1}{G} \sum_{g=1}^{G} P\left( \pmb{s}_t 
\left| \pmb{s}_0; \bm{\theta}^{(g)}, \bm{w}^{(g)} \right. \right).
\]
In addition, given the true regulatory mechanistic model $\mathbf{v}(\pmb{s}_t; \pmb{\theta}^c)$, we can construct the predictive distribution $P(\pmb{s}_t \mid \pmb{s}_0;\bm{\theta}^c)$ to assess the prediction performance by using different Bayesian model inference approaches. 

We evaluate the performance of the posterior predictive distribution obtained by MALA with and without adjoint sensitivity, as well as ABC, using the Kolmogorov–Smirnov (K-S) statistic. The K-S statistic quantifies the maximum discrepancy between the posterior predictive distribution and the predictive distribution of the true model. {Specifically, it is defined as
$
D = \sup_{i} | F(s_t^{i}\mid \pmb{s}_0;\bm{\theta}^c) - F(s_t^{i} \mid \pmb{s}_0,\mathcal{D}_m)|$ for $i = 1,2, \ldots, p,
$
where \( F(s_t^{i} \mid \pmb{s}_0;\bm{\theta}^c)\) and \( F(s_t^{i} \mid \pmb{s}_0,\mathcal{D}_m)\) are the empirical cumulative distribution functions (CDF)  of the $i$-th component of state $\pmb{s}_t$ derived from samples of the true model’s predictive distribution and the posterior predictive distribution, respectively.} 
\textit{The results in Table~\ref{tab:mala-abc-comparison} show MALA with adjoint sensitivity analysis has smaller K-S distance and demonstrates superior predictive performance for all key states when the number of batches is $m=3,5$.} 
The results are based on $R=30$ macro-replications.
In each $r$-th macro-replication with $r=1,2,\ldots,R$, we use $G=2000$ samples to construct the empirical distributions. We report the 95\% confidence intervals of the K-S distances for the predictions of the three key states at $t=72$ hour, computed as
$
\bar{D} \pm 1.96 \times \frac{S_D}{\sqrt{R}}$,  where $
\bar{D} = \frac{1}{R} \sum_{r=1}^{R} D^{(r)}$ and $S_D = \left[ \frac{1}{R-1} \sum_{r=1}^{R} \left(D^{(r)} - \bar{D} \right)^2 \right]^{1/2}.$

\begin{figure}[h]
    \centering
    \includegraphics[width=\textwidth]{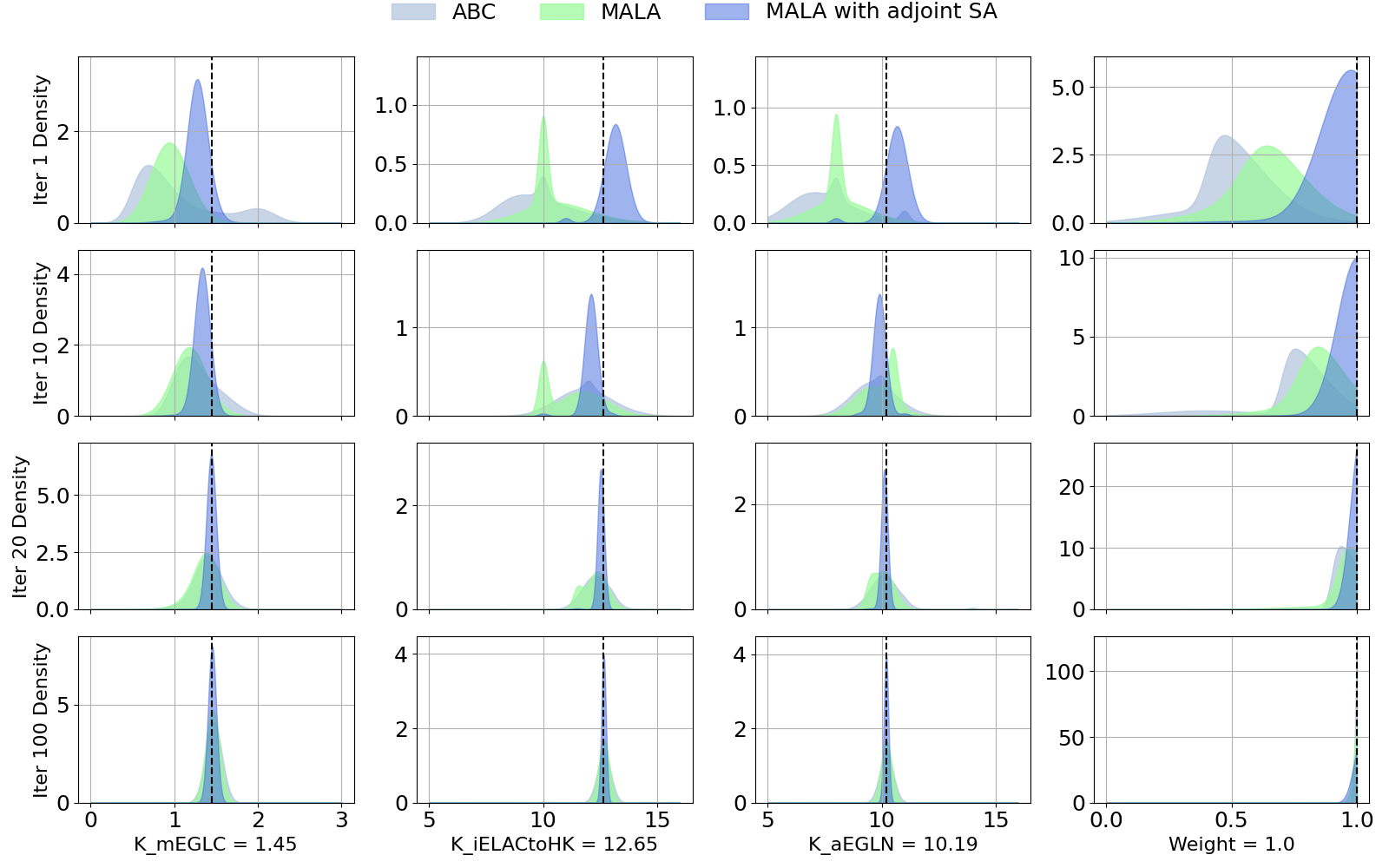} 
    \caption{Asymptotic consistency of the posterior distributions of $\bm{\theta}$ and $\bm{w}$ across iterations.}
    \label{fig:ABC_param}
\end{figure}

The performance improvement is further illustrated in the estimated posterior distributions of $\bm{\theta}$ and $\bm{w}$ as shown in Figure~\ref{fig:ABC_param}. Here we select one representative parameter for each regulatory mechanism: $K_{m,\text{EGLC}}$ is the dissociation constant in Reaction 2; $K_{i,\text{ELACtoHK}}$ captures the non-competitive inhibition of ELAC on EGLC uptake in Reaction 2; and $K_{a,\text{EGLN}}$ reflects the allosteric activation effect of EGLN on ELAC production in Reaction 3. 
The results show the MALA with adjoint SA produces posterior distributions with higher concentration—defined as posterior mass surrounding the true parameter value $\bm{\theta}^c$—and achieves faster convergence toward $\bm{\theta}^c$. In addition, as shown in the last column of Figure~\ref{fig:ABC_param}, the weight $w_{k^*}$ of the true model, corresponding to the correct combination of regulatory modules R1–R4 in Figure~\ref{fig:MetaNetwork}, also converges to 1 more rapidly compared to other state-of-the-art methods.

\section{Conclusion}
\label{sec: conclusion}
In this paper, we consider mechanistic models of biomanufacturing process in the form of SDEs. We present a novel symbolic and statistical learning framework that leverages on the existing knowledge of bioprocess mechanisms and sample efficiently recovers bioprocessing regulatory mechanisms. By constructing a mixture model of candidate mechanisms and employing Bayesian inference with MALA, the framework enables joint learning of model structure and kinetic parameters while quantifying uncertainty. 
In addition, 
adjoint sensitivity analysis is integrated into MALA, that can quickly estimate the initial bias, reduce the warmup time of posterior sampling, and accelerate the convergence. Empirical results demonstrate that the proposed framework outperforms the state-of-the-art Bayesian sampling approaches, including MALA and ABC, in terms of prediction accuracy and recovery of underlying regulatory mechanisms. The study highlights the importance of interpretable and sample-efficient learning strategies for digital twin development in biomanufacturing.

\section*{Acknowledgement}
We gratefully acknowledge funding support from National Science Foundation Grant CAREER CMMI-2442970 and National Institute of Standards and Technology Grant 70NANB24H293 to Dr. Wei Xie.

\footnotesize

\bibliography{sn}

\end{document}